\documentclass[a4paper]{article}
\usepackage{arxiv}

\usepackage[utf8]{inputenc} 
\usepackage[T1]{fontenc}    
\usepackage{hyperref}       
\usepackage{url}            
\usepackage{booktabs}       
\usepackage{amsfonts}       
\usepackage{amsmath}
\usepackage{nicefrac}       
\usepackage{microtype}      
\usepackage{svg} 
\usepackage{graphicx}
\usepackage{subcaption}
\usepackage[numbers]{natbib}
\usepackage{doi}

\newcommand{\AuthorEntry}[3]{%
  \begin{minipage}[t]{0.3\textwidth}
    \centering
    \textbf{#1}\textsuperscript{#2}\\
    \texttt{#3}
  \end{minipage}
}

\newcommand{\AffilEntry}[2]{%
  \textrm{\textsuperscript{#1} #2}
}

\title{VIBE: \textbf{V}ideo‑\textbf{I}nput \textbf{B}rain \textbf{E}ncoder for \textnormal{f}MRI Response Modeling}

\author{
\AuthorEntry{Daniel Carlström Schad}{*1}{schad@cbs.mpg.de}\And
\AuthorEntry{Shrey Dixit}{*1, 2}{dixit@cbs.mpg.de}\And
\AuthorEntry{Janis Keck}{*1, 3, 5}{janis.keck@maxplanckschools.de}\AND
\AuthorEntry{Viktor Studenyak}{*1, 5}{studenyak@cbs.mpg.de}\And
\AuthorEntry{Aleksandr Shpilevoi}{1}{ashpilevoii@gmail.com}\And
\AuthorEntry{Andrej Bicanski}{1, 4}{bicanski@cbs.mpg.de}
}

\affiliations{
\AffilEntry{*}{Equal contribution}\\
\AffilEntry{1}{Max Planck Institute for Human Cognitive and Brain Sciences, Leipzig, Germany}\\
\AffilEntry{2}{Institute of Computational Neuroscience, University Medical Center Hamburg-Eppendorf, Hamburg, Germany}\\
\AffilEntry{3}{Max Planck Institute for Mathematics in the Sciences, Leipzig, Germany}\\
\AffilEntry{4}{ScaDS.AI - Center for Scalable Data Analytics and Artificial Intelligence, Leipzig, Germany}\\
\AffilEntry{5}{Max Planck School of Cognition, Leipzig, Germany}
}

\renewcommand{\headeright}{}
\renewcommand{\undertitle}{}
\renewcommand{\shorttitle}{VIBE: Video‑Input Brain Encoder}

\begin{document}
\maketitle

\begin{abstract}
We present VIBE, a two‑stage Transformer that fuses multi-modal video, audio, and text features to predict fMRI activity. Representations from open‑source models (Qwen2.5, BEATs, Whisper, SlowFast, V‑JEPA) are merged by a modality‑fusion transformer and temporally decoded by a prediction transformer with rotary embeddings. Trained on 65 hours of movie data from the CNeuroMod dataset and ensembled across 20 seeds, VIBE attains mean parcel‑wise Pearson correlations of {0.3225} on in‑distribution Friends S07 and {0.2125} on six out‑of‑distribution films. An earlier iteration of the same architecture obtained 0.3198 and 0.2096, respectively, winning Phase-1 and placing second overall in the Algonauts 2025 Challenge.
\end{abstract}


\section{Introduction}

Humans process a constant stream of sensory information, with different sensory modalities processed first by primary sensory cortices and then combined in polymodal association cortices to create integrated representations of experience. This feed-forward stream is accompanied by feedback connections across cortex, as well as interactions with sub-cortical areas. Although our understanding of stimulus-driven neural responses in classical experimental paradigms has improved greatly \cite{yamins_using_2016, stigliani_encoding_2017}, it is still unclear how this transfers to the complex situations encountered in the real world. Compared to static stimuli, video input approximates the constant stream of sensory information in naturalistic settings. To accurately predict brain activity under these conditions, it is crucial to utilize the features derived from all available sensory modalities. Encoding models have become increasingly relevant as a way of studying neural representations under complex naturalistic conditions \cite{van_gerven_primer_2017}. Addressing the challenge of mapping multi-modal video data to corresponding fMRI activity, the Algonauts 2025 Challenge \cite{gifford_algonauts_2025} utilized the CNeuroMod dataset \cite{boyle_courtois_2020}. This dataset consists of extensive recordings (65 hours) of brain activity collected while participants watched various TV series and movies.

To effectively model the relationship between multi-modal data and fMRI BOLD responses, we developed VIBE (Video‑Input-to‑Brain Encoder), an architecture composed of two components: a modality fusion transformer and a prediction transformer. The modality fusion transformer integrates features extracted from distinct sensory modalities, while the prediction transformer establishes temporal correspondences, aligning current brain activity with past activity states.

During the initial training phase of the Algonauts competition on in-distribution data, VIBE achieved a mean correlation of $0.3198$, securing first place. In the subsequent model selection phase, which involved testing on out-of-distribution data, it achieved a mean correlation of $0.2096$, placing second overall in the challenge.

After the competition concluded, we revisited our codebase, identified minor implementation bugs, and back‑ported the most effective Phase‑2 techniques into the Phase‑1 pipeline. As a result, both leaderboard scores increased slightly: the in‑distribution mean correlation rose from $0.3198$ to $0.3225$, and the out‑of‑distribution score climbed from $0.2096$ to $0.2125$. Unless otherwise noted, all performance figures reported in the remainder of this paper refer to these post‑challenge results.

\section{Data}
Training data was provided by the organizers of the Algonauts competition. The training set, a part of the larger CNeuroMod dataset \cite{boyle_courtois_2020}, included seasons 1-6 of \textit{Friends} and five films: \textit{The Wolf of Wall Street}, \textit{Life}, \textit{Hidden Figures}, and \textit{The Bourne Supremacy}, accompanied by corresponding whole-brain fMRI data from 4 out of 5 subjects (one excluded, see below). The stimuli included video clips and corresponding transcripts in English.

The fMRI data used in the competition were provided in pre-processed format in the MNI152 space \cite{brett_problem_2002} and assigned into 1000 distinct parcels using the Schaefer atlas \cite{schaefer_local-global_2017}. Subject 4 was excluded from the dataset due to issues with data quality. In the initial training phase, models were trained on the provided training set and tested on in-distribution data from season 7 of \textit{Friends}. In the subsequent model selection phase, models were tested for generalization on an out-of-distribution dataset featuring clips from six different films/TV Shows: \textit{The Pawnshop}, \textit{World of Tomorrow}, \textit{Princess Mononoke}, \textit{Planet Earth}, and \textit{Passe-partout}. A more thorough description of the dataset and pre-processing steps can be found in \citeauthor{gifford_algonauts_2025} and the corresponding GitHub repository.

\section{Methods}
The methods are divided into two subsections. First, we describe the features extracted from video files and their accompanying transcripts. This is followed by a detailed explanation of the VIBE architecture.
\subsection{Feature Extraction}
Feature extraction plays a critical role in achieving a high correlation between predicted and actual fMRI responses. Since different regions of the brain are responsible for different functions, it is important to extract features using models that align with the most current hypotheses concerning the functions of these regions. This task-specific alignment increases the chances that the extracted representations are related to the brain areas being modeled.

The stimuli consist of video clips, each approximately 10 minutes in duration. Since the temporal order of frames in long-running video clips contains significant information, we chose feature extraction models that are sensitive to this temporal information across all modalities—audio, video, and text. This led us to exclude models that operate on single images without any temporal context from consideration.

The models we used for feature extraction are listed below in order of their importance to the final performance.
\subsubsection{Text Features: Qwen2.5 14B}
Text features were extracted from the transcripts using the Qwen2.5 14B model \cite{qwen_qwen25_2025}. This is a large language model with 14.7 billion parameters, trained on 18 trillion tokens in more than 29 languages.

Thanks to the model's large context window of 128,000 tokens, we were able to input the entire transcript of each clip at once. In order to extract textual features, we passed the full transcript into the model and computed the average of the activations from the last four hidden layers for each token. These averaged token representations were then aligned with the fMRI time resolution (1.49 seconds). If multiple tokens occurred within a single repetition time (TR), we took the mean of their representations to form the final feature vector for that TR.

Along with the main transcript, we also provided a manually created one-line self-written description of each clip, which typically mentioned the movie's name, duration, and occasionally the director's name. We also added brief instructions asking the model to focus on emotional content in the transcript. Since no transcripts were provided for the Charlie Chaplin movie, whitespace was used.

We observed a significant increase in prediction accuracy when using the full transcript compared to when using a sliding window around the current TR. This performance gain was mostly realized in the default mode network. This is likely because it allowed the model to capture long-term dependencies such as scene transitions, long running emotional states, and aspects of the narrative structure. 


\subsubsection{Visual Features: V-JEPA 2}
Further, we used the features from the model based on the Joint-Embedding Predictive Architecture model, V-JEPA 2, with the ViT-L/16 backbone and action classification head trained on a broad range of video stimuli from datasets representing different actions and contexts \cite{assran_v-jepa_2025}. The model learns to make predictions of the actions from contextual cues for each video input. To extract the features for the challenge dataset we used a context window of 6s (3s before the TR and 1.51s after the TR) as the input at time $t$. Hidden states from the layers were spatially downsampled using adaptive average pooling, reducing the grid patches from 16x16 to 3x3. Finally, the [T x 3 x 3 x D] tensor was averaged over T dimension and flattened.

\subsubsection{Audio Features: BEATs}
Audio features were extracted using BEATs (Bidirectional Encoder representation from Audio Transformers), a self‑supervised audio model that learns acoustic representations by iteratively training an acoustic tokenizer and a Transformer encoder \cite{chen_beats_2023}. We collected the hidden activations from the top third of BEATs Transformer layers and averaged them across layers to obtain a single representation per time frame. Because BEATs was pre‑trained on AudioSet, where examples are 10‑second clips, we treated 10 seconds as the model’s effective context length. 

Processing only the 1.49‑second interval aligned to a single fMRI TR would provide too little temporal context, especially for complex naturalistic stimuli. We therefore used an overlapping sliding‑window procedure: 10‑second audio windows advanced in 1.49‑second strides (the TR duration). Within each 10‑second window, we retained activations only for the final 1.49‑second segment corresponding to the current TR. We then computed both the mean and the standard deviation of these activations to preserve limited within‑TR temporal variation while producing a single feature vector per TR.

\subsubsection{Audio Features: Whisper-V3}
We also extracted phonetically sensitive audio features using Whisper V3 \cite{radford_robust_2022}, a speech recognition model from OpenAI whose encoder representations have been shown to encode sub‑word (phonetic) distinctions that can be used for downstream phoneme and prosody prediction tasks, even though the model natively produces text rather than explicit phoneme labels \cite{goldstein_unified_2025}.

Feature extraction followed the same procedure as that used for BEATs. We averaged activations across the top third of the Whisper encoder layers within 10‑second sliding windows advanced in 1.49‑second strides (the TR). For each window, we retained activations corresponding to the final 1.49 seconds, then computed the mean and standard deviation to yield one feature vector per TR.

\subsubsection{Visual Features: SlowFast R101}
We used the SlowFast R101 video model, which processes video through two parallel pathways: a slow pathway that samples frames sparsely to emphasize spatial semantics, and a fast pathway that samples more densely to capture motion at higher temporal resolution \cite{feichtenhofer_slowfast_2019}. At the top of the network, features from the two pathways are pooled and concatenated to form a joint representation that can be used for downstream tasks. 

Because the model operates on short fixed‑length clips, we processed each fMRI TR by sampling 32 frames corresponding to that TR and feeding those frames as a clip to the model. This clip length reflects the limited temporal context the model can take at once in our setup.

We extracted activations from multiple internal layers of the network. Among these, the pooled concatenated representation (\texttt{pool\_concat}; the fusion point where Slow and Fast pathway features are joined) and an intermediate Slow‑pathway activation, which we refer to as \texttt{slow\_act\_3}, produced features that were the most helpful when combined with the other features used in the model.

\subsubsection{Omni Features: Qwen2.5 Omni 3B}
Qwen2.5 Omni 3B is a multi-modal (“omni”) model that can process video, audio, and text inputs and generate text or speech outputs \cite{xu_qwen25-omni_2025}. For our purposes, we passed only the video frames and the accompanying audio track from each clip into the model to extract features; no textual input was used.

Extracting representations from its internal layers revealed that features from a single layer (24th layer from the \texttt{thinker} tower) alone were strongly predictive of fMRI responses. Using just that layer yielded an average correlation of about 0.26 on the Friends S07 dataset, which would have placed us in roughly the top ten teams in the competition after ensembling (see below). However, when these omni features were combined with the other feature sets described above, the added benefit was insignificant for S07 but marginally better for out-of-distribution movies.

\subsubsection{Text features: LaBSE}
To improve out-of-distribution prediction performance for Passe-partout, a language-agnostic language embedding was extracted from the French transcripts. To this end, we used a language-agnostic BERT sentence embedding model \citep{feng_language-agnostic_2020} to extract features that were invariant to transcript language. Predictions for Passepartout thus used the same model as the other OOD dataset movies, with the addition of LaBSE features extracted from each TR from layers 8-10. These features correlate more strongly with semantic information across languages than earlier layers, but still contain enough information to be predictive. 

\subsection{Model}
\begin{figure}[ht]
    \centering
    \includegraphics[width=0.9\linewidth]{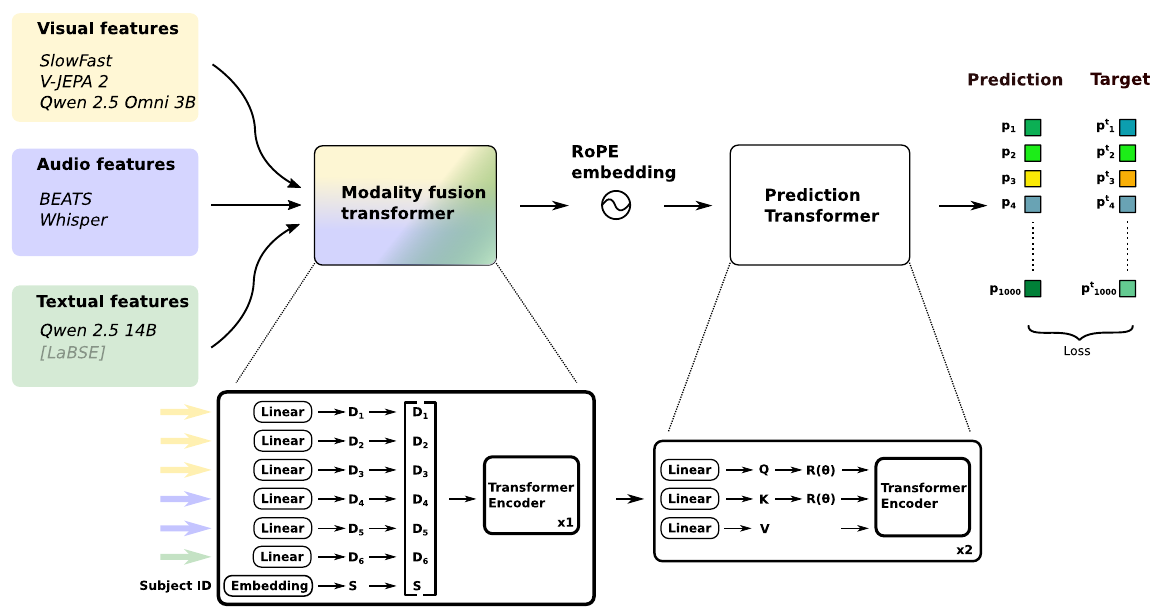}
    \caption{Overview of the VIBE architecture. (1) We extract a diverse set of visual features from different layers of the SlowFast, V-JEPA 2, and Qwen 2.5 Omni 3B models. For audio, we employ both the BEATS and Whisper models to obtain complementary auditory features. Textual features are derived using the Qwen 2.5 14B model. (2) The extracted features are linearly projected, concatenated, and subsequently integrated using a modality fusion transformer. Rotary positional embeddings are then applied to the query and key vectors of the fusion transformer output, followed by further processing with a transformer encoder.}
    \label{fig:model}
\end{figure}

An overview of VIBE's architecture is provided in figure~\ref{fig:model}. The core structure is based on a transformer architecture and is divided into two main components: the \textbf{Modality Fusion Transformer} and the \textbf{Prediction Transformer}.

The first component, the \textbf{Modality Fusion Transformer}, is responsible for computing cross-attention across modalities. Before fusion, each feature is passed through a projection layer—a simple feedforward layer—that reduces its dimensionality to 256. In addition to these projected features, we included a 256-dimensional subject embedding. An alternative approach would have been to train a separate model for each subject, but we hypothesized that using a single shared model with subject embeddings would generalize better. This assumption was not explicitly tested due to the time constraints given by the competition.

The Modality Fusion Transformer consists of a single layer. Importantly, it does not model temporal dependencies. Instead, it performs cross-attention independently for each time point (TR), meaning that features at a given TR are fused only with features from the same TR across modalities. The outputs of this transformer, one fused representation per feature per TR, are then concatenated into a single vector for the next stage.

The concatenated features are then passed to the second component: the \textbf{Prediction Transformer}. This module models the temporal dependencies across TRs within each clip, similar to how large language models process sequences of tokens. The Prediction Transformer uses two layers, and its output representations are passed through a final feed-forward layer to predict the fMRI responses.

Initially, we experimented with fixed sinusoidal positional embeddings to make the model time aware. However, these embeddings degraded performance, likely due to interference with feature representations. Concatenating them also failed to improve results, possibly because it made learning temporal dependencies harder for the model. To resolve this, we adopted Rotary Positional Embeddings (RoPE), which encode relative positional information through rotations in the embedding space \cite{su_roformer_2023}. Unlike absolute embeddings, RoPE enables the model to better capture the relative positions between tokens (or TRs). This change significantly improved model performance.

\subsubsection{Loss}
The training objective combined a correlation loss with a small Mean‑Squared Error (MSE) term:
\[
\mathcal{L}
\;=\;
\mathcal{L}_{\mathrm{Pearson}}
\;+\;
\lambda_{\mathrm{MSE}}\,
\mathcal{L}_{\mathrm{MSE}},
\qquad
\lambda_{\mathrm{MSE}} = 0.03.
\]
A lightly weighted MSE term anchors the output scale and offset to the scale of the data, stabilizing optimization without sacrificing correlation.

We trained with the AdamW optimizer, which provides adaptive moments while using decoupled $\ell_2$ regularization. Spatial, network Laplacian, and additional weight-decay penalties were explored in development but proved negligible for validation performance and are therefore not discussed in this report.

\subsubsection{Backwards causality}
Most Transformer-based language models are trained using a causal mask, which prevents the model from attending to future tokens during training. However, according to the predictive coding theory, humans not only attend to past experiences, but also predict future occurrences \citep{rao_predictive_1999, friston_predictive_2009}. Therefore we hypothesized that if the subjects predicted future experiences when watching the movies, removing the causal mask from the transformer could improve performance. We found that removing the causal mask (thus allowing the model to attend to future time points) led to an improvement in Pearson correlation by approximately 0.002. This suggests that future stimulus information contributes to predicting the current brain response.

\subsubsection{Hemodynamic Response Function}
In neuro-imaging experimental setups, it is standard to assume a canonical \textit{hemodynamic response function} (HRF) that models the delayed response of fMRI signal to a stimulus due to the slowness of the vascular/metabolic adjustments that ultimately cause the signal. In practice, this is achieved by convolving time-series of interest, for example of predictors, with a function modeling the HRF \cite{lindquist_modeling_2009}.

We attempted using this as an inductive bias for our model, by either convolving final predictions with a hand-crafted HRF, or using an additional learnable 1-d convolution of our output activity time-series. Both approaches decreased performance. We concluded that the HRF, which represents an idealized estimation of the BOLD response, is too inflexible and that the model is better off learning a nuanced slow temporal response through its internal temporal context.

\subsubsection{Ensembling}
Ensembling proved to be the single most effective lever for boosting performance. We averaged the outputs of 20 independently trained models; beyond this size, additional members yielded only marginal gains. Every model in the ensemble shared the same architecture and parameter count, differing solely in random weight initialization and the order in which training mini‑batches were presented.

We also experimented with more sophisticated aggregation schemes---including correlation‑weighted averaging, leave‑one-out stacking, and gradient‑boosting---but none of these approaches outperformed the mean of 20 seeds. The simple ensemble therefore offered the best trade‑off between accuracy and implementation complexity.

\subsubsection{Functional network specialization}\label{sec:vis_model}
Each of the 1000 parcels of FMRI data may be uniquely assigned to one of 7 non-overlapping functional networks \cite{thomas_yeo_organization_2011}. In brief, these networks are defined by first taking high correlation of parcel activity time-series in the resting state as a proxy for anatomical connectivity, secondly constructing a connectivity profile for each parcel by stacking these correlations together, and thirdly applying unsupervised clustering on the connectivity profiles. Thus, parcels in each of the 7 networks share a similar connectivity structure. The seven networks identified are: Visual, Somatomotor (SomMot), Salient Ventral Attention (SalVentAttn), Dorsal Attention (DorsAttn), Frontoparietal Control (Cont), Default Mode Network (Default), and Limbic.

We hypothesized that constructing network-specific models might facilitate learning, as neural activity in the networks is highly correlated \textit{a priori}. This could make it easier for the model to build representations useful for their prediction; conversely, when including separate, unrelated networks the model might be forced to first disentangle these, hindering learning.

To train the network specific models, we applied a parcel mask to the loss function $\mathcal{L}^{y_i}$ such that only parcels in the corresponding network $y_i\in \mathbb{Y}_{Visual}\subset \mathbb{Y}$ contribute to the loss. After some experimentation, we then decided to use three models in total
$m_{Visual}, m_{Default},m_{All}$ trained on the Visual, Default and all networks respectively. We then combined their predictions by taking the predictions for Visual, Default from the first two models, and for all other parcels from the last. 

\begin{figure}[htbp]
    \centering
\includegraphics[width=0.45\linewidth]{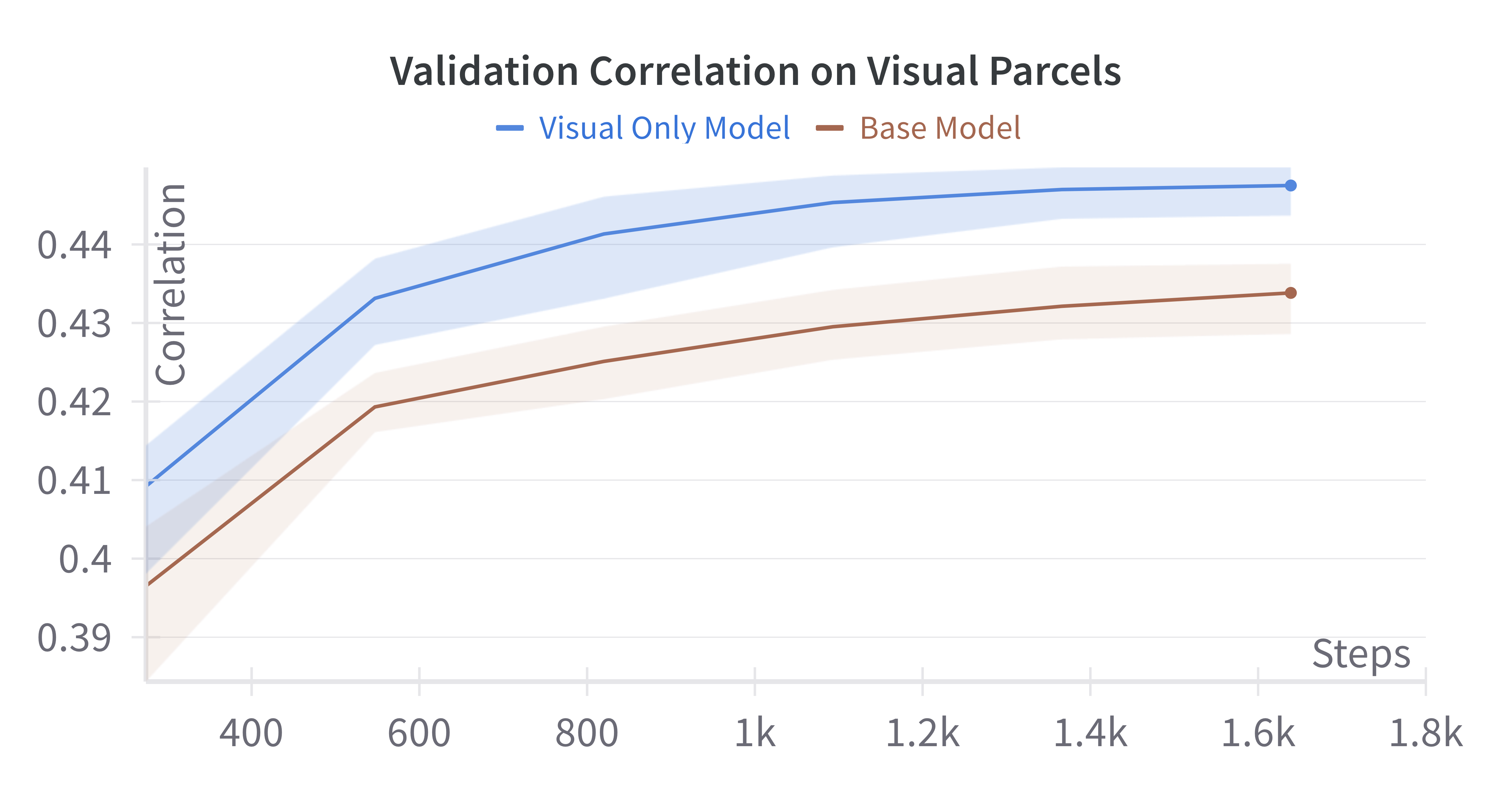}
\includegraphics[width=0.45\linewidth]{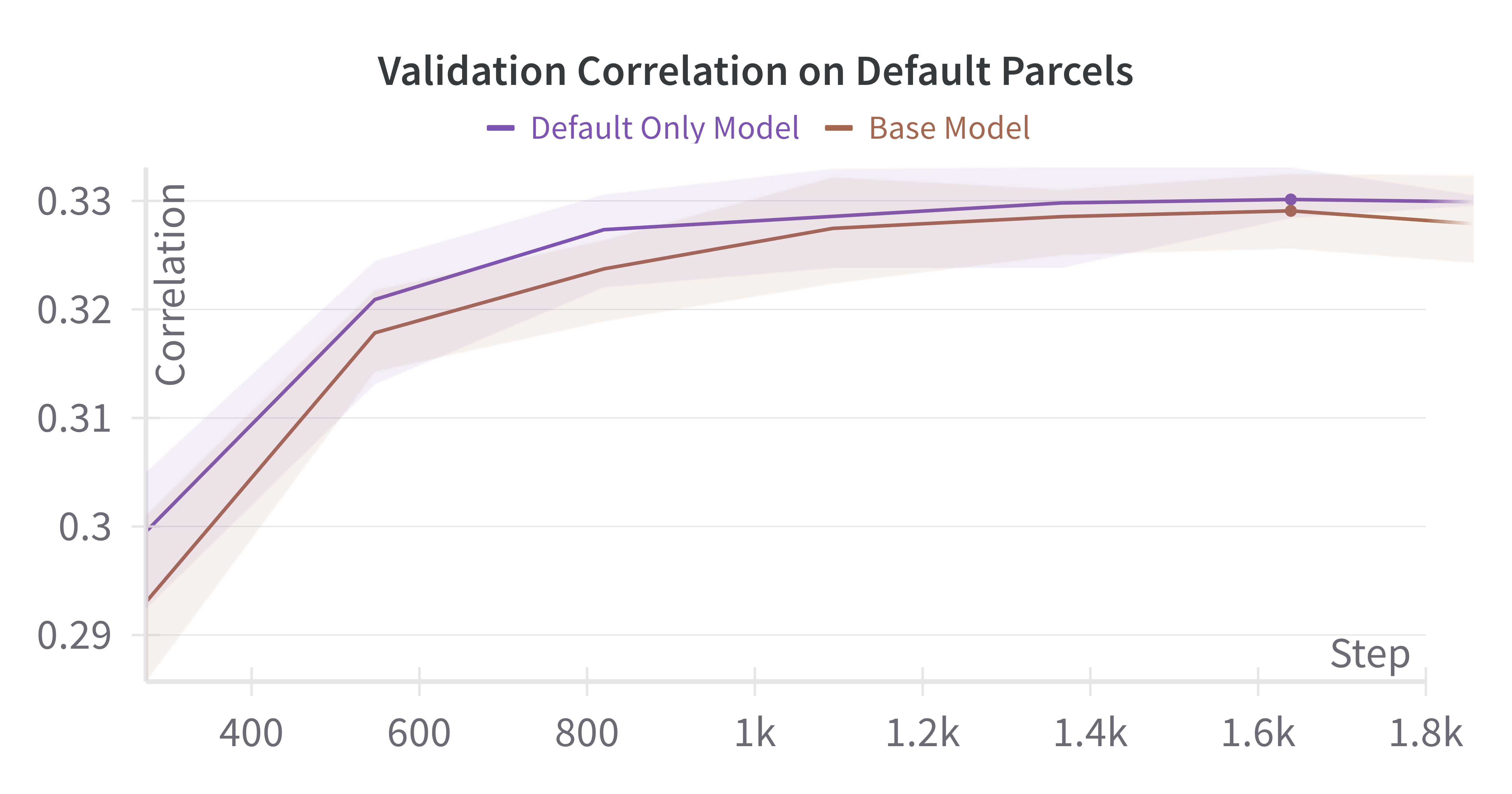}
\includegraphics[width=0.45\linewidth]{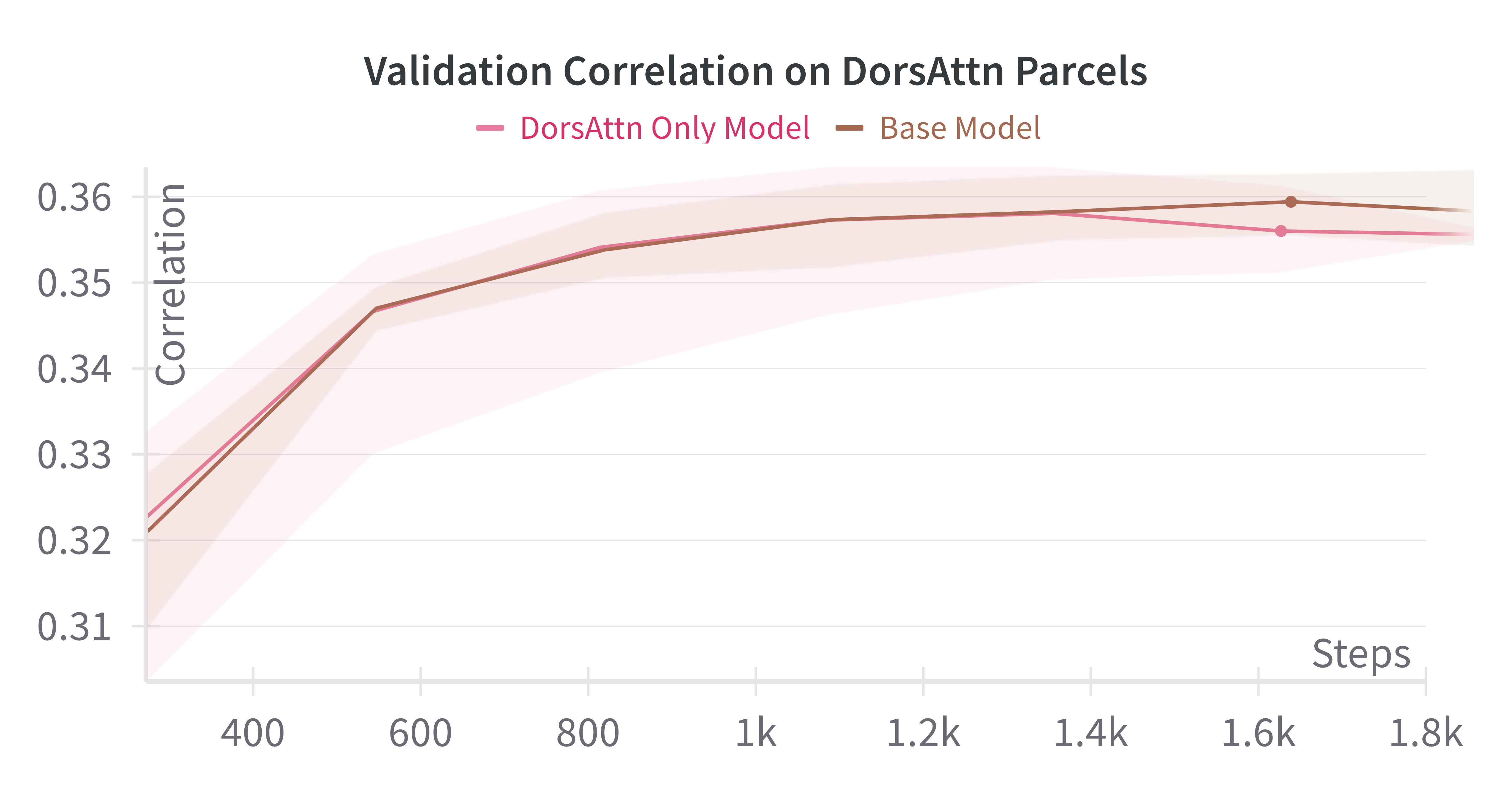}    
    \caption{Validation scores for models separately trained on subsets of parcels corresponding to separate functional networks. Visual model outperforms full-cortex baseline; Default shows moderate gains; other networks show no improvement.}
    \label{fig:roi_models}
\end{figure}

\subsubsection{Training details}
The model was initially trained with season six of friends held out for validation. 
Models were trained for at most 15 epochs with a batch size of 4, using AdamW ($\beta_1{=}0.9,\beta_2{=}0.999$, weight-decay 0.0001). The learning rate was set to $1{\times}10^{-4}$, using a cosine schedule. Gradients were clipped at a $\ell_2$ norm of 5.0.
Early stopping monitored validation Pearson $r$ with a patience of 2 epochs; the best checkpoint was saved.
After training, the model was retrained on the full dataset until the early stopping epoch count from initial training.
For submissions, the results of 20 training runs with random initial seeds were averaged, and the predictions from separate Visual and Default ensembles were used for corresponding parcels.
Training a single model took 45 GPU-minutes on one NVIDIA A100-40 GB (Pytorch 2.7, CUDA 12.6); peak memory was 19 GB. Reproducing the full model requires 60 GPU hours and consumes approximately 1 TB of storage on disk.
Parameters reproducing these settings can be found under \texttt{configs/} in the GitHub repository.

\section{Results}
VIBE yields consistent gains over the organizer's benchmark ridge regression model. Averaged across the 1000 cortical parcels, our 20-seed ensemble reaches a mean Pearson $r$ of $0.3225$ on the in-distribution Friends S07 split and $0.2125$ on the six out-of-distribution films—results that placed us first in Phase 1 and second overall in Phase 2 of the Algonauts 2025 challenge.

\subsection{Predictive Performance on In‑ and Out‑of‑Distribution Data}

\begin{figure}[htbp]
    \centering
    \begin{subfigure}[t]{.45\textwidth}
        \centering
        \includegraphics[width=\linewidth]{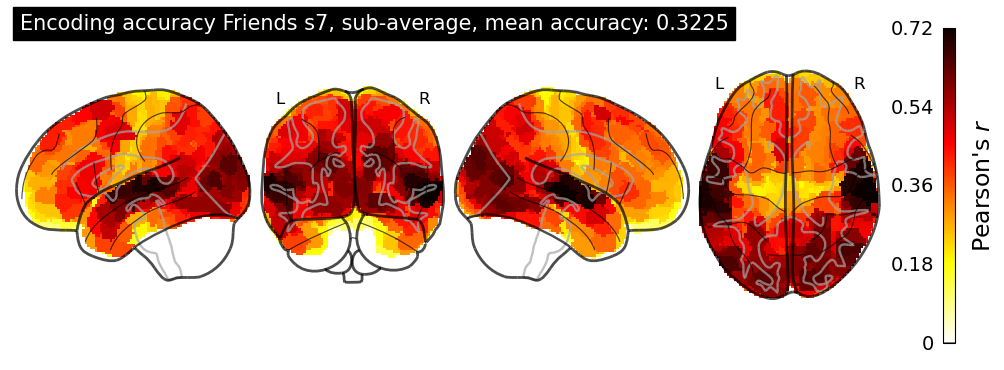}
        \caption{Model score for in-distribution testing on season 7 of Friends.}\label{fig:scores_a}
    \end{subfigure}
    \begin{subfigure}[t]{.45\textwidth}
        \centering
        \includegraphics[width=\linewidth]{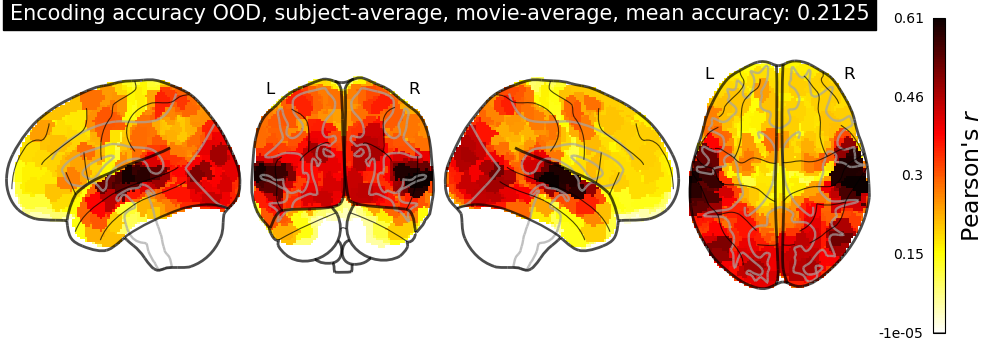}
        \caption{Model score for out-of-distribution test set movies.}\label{fig:scores_b}
    \end{subfigure}
    \caption{Model scores for in and out of distribution test sets.}
    \label{fig:scores}
\end{figure}

For in-distribution testing on Season 7 of Friends, the model was trained using Season 6 of Friends for early stopping during an initial training phase before being retrained on the full dataset. Detailed descriptions of model structure and parameter choices can be found on GitHub. Overall correlation score on the in-distribution test set was $0.3225$, see figure \ref{fig:scores_a}. As a comparison, the baseline ridge model achieved $0.2033$ correlation for the in-distribution evaluation \citep{gifford_algonauts_2025}.

For out of distribution movies, we used the same model that was developed during phase 1, with minor modifications. French language features were included only in predictions of Passe-partout fMRI BOLD responses. The average correlation with out-of-distribution movies was $0.2125$. The baseline ridge model had a $0.090$ correlation with fMRI BOLD responses \citep{gifford_algonauts_2025}. Overall, our model outperformed the baseline model by $0.11925$ and $0.12299$ for in-distribution and out-of-distribution tests, respectively.

\begin{table}[ht]
    \centering
    \caption{Performance comparison vs. baseline.}
    \begin{tabular}{lcc}
        \toprule
        Model & Friends S07 ($r$) & OOD ($r$) \\
        \midrule
        Official baseline \citep{gifford_algonauts_2025} & 0.20328 & 0.08952 \\
        VIBE & 0.32253& 0.21251\\
        \midrule
        VIBE vs. baseline & +0.11925& +0.12299\\
        \bottomrule
    \end{tabular}
    \label{tab:perf}
\end{table}

Performance improvements were cumulative across three key modifications. Ensembling across 20 random seeds improved mean parcel-wise correlation by approximately 0.011-0.012 over single-seed performance. Adding dedicated models for the visual cortex and default mode network (see Section \ref{sec:vis_model}) contributed a further 0.003–0.004, while removing the causal mask from the prediction transformer yielded an additional 0.002 percentage points. These effects were largely additive, resulting in a final ensemble score of 0.3225 on Friends S07.

The introduction of separate visual and default mode network models increased model performance notably. The largest performance improvement was realized in visual cortex. This may be because parcels in visual cortex exhibit distinct stimulus-driven responses that may not be captured as well when projections for visual features are also trained to be informative for other brain regions. 

Interestingly, removing the causal mask from the model improved performance. This finding aligns with the theory of predictive coding, suggesting that the brain constantly builds expectations about future observations based on current and past observations \cite{rao_predictive_1999, friston_predictive_2009}. The results further support the idea that the brain may begin responding to stimuli in anticipation of future experience.

\subsection{Parcel-wise Feature Contribution Mapping}
An interesting question to ask is: which features are predictive of which regions of the brain? Understanding this mapping could help generate new hypotheses about the functional roles of different brain regions, which could then be tested in controlled experiments.

\begin{figure}[ht]
    \centering
    \includegraphics[width=0.9\linewidth]{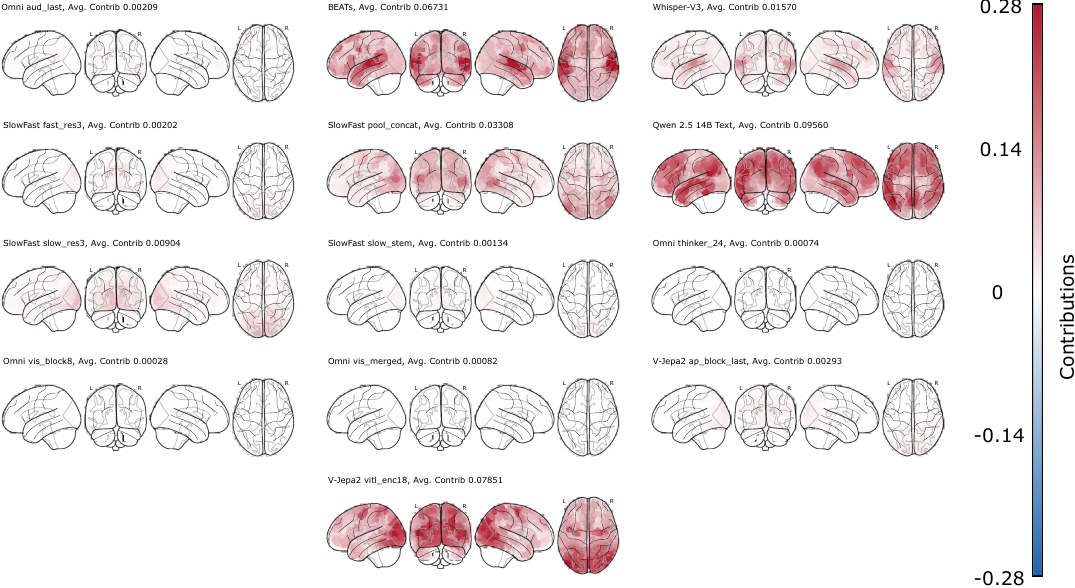}
    \caption{Multiperturbation Shapley‑Value Analysis (MSA) of parcel‑wise feature contributions. Warm colors mark parcels where a feature set improves prediction accuracy the most. Text features from Qwen 2.5 14B dominate higher‑order language and default‑mode areas as well as parietal attention areas, BEATs and Whisper peaks align with auditory cortex and speech parcels, and SlowFast/V‑JEPA highlight occipital and motion‑sensitive regions. Omni‑model features contribute negligibly across the cortex.}
    \label{fig:msa}
\end{figure}

To answer this, we applied Multiperturbation Shapley-Value Analysis (MSA) \cite{dixit_who_2025, fakhar_msa_2021, shapley_value_1953}—a method that systematically perturbs feature vectors to estimate their contribution to the model’s output. We investigated how these perturbations affected parcel-wise correlation performance. This allowed us to estimate the contribution of different features for each brain parcel. The results of this analysis are summarized in Figure~\ref{fig:msa}.

We observed a pronounced hierarchy in how each feature set drives parcel‑wise prediction accuracy. Text representations from Qwen 2.5 14B stand out with the highest mean contribution ($\approx 0.096$), followed by mid‑level visual embeddings from V‑JEPA‑2 (\texttt{vitl\_enc18}) at $\approx 0.079$ and acoustic features from BEATs at $\approx 0.067$. SlowFast's pooled visual representation contributes more modestly ($\approx 0.033$), while Whisper‑V3's phonetic layer adds a small but non‑negligible boost ($\approx 0.016$). By contrast, all tested omni features register near‑zero average contributions---e.g. \texttt{aud\_last} at $\approx 0.002$ and \texttt{thinker\_24} below $0.001$---indicating that they add little to Friends S06 predictions once the modality‑specific encoders are present.

Spatially, the contribution maps follow expected neuro-anatomical patterns.  Text features dominate parcels in the frontal cortex, posterior cingulate, and left temporal regions associated with the default‑mode and language networks. BEATs contributes most to bilateral superior temporal parcels around auditory cortex, while Whisper also contributes mostly to auditory regions.  Visual encoders split their influence: SlowFast and V‑JEPA both highlight occipital areas and motion‑sensitive dorsal stream parcels, with V‑JEPA extending further into visual cortex. Taken together, these results underscore that modality‑specific representations provide complementary, anatomically plausible signals, whereas the omni features offer little unique predictive power.

\section{Conclusion}
Our approach combines meaningful multi-modal features with a temporally sensitive architecture that fuses these features together to predict brain activity. We identify several key ingredients for the high prediction capabilities of our model: the features obtained using domain-specific open-source models, RoPE embeddings for relative positioning of the tokens, ensembling of the models, and specialized models for the visual and default-mode networks. Using Shapley-Value Analysis we were able to identify which features are most helpful in predicting each parcel.

Although the model was developed under tight time constraints, with the Algonauts challenge as the primary goal, two interesting scientific observations can be made. First, backwards causality improved model performance to a small extent. The most straight-forward interpretation is that this reflects some predictive coding aspects of brain activity. However, the limited magnitude of the contribution simultaneously suggests limits to predictive capabilities. Second, some brain areas were more poorly predicted than others. While a complete breakdown by regions is out of scope here, among the most poorly predicted areas (in phase 1) we find the intraparietal sulcus (part of the dorsal attention network), the temporal pole (associated with several high-level cognitive processes), limbic areas and parts of the ventral attention network (involved, among other functions, in social behavior and emotional processing). In phase 2 correlations dropped substantially in frontal and inferior temporal regions. Poor performance on some areas could reflect hard limits, as the activity of some areas may not be derivable from the stimuli themselves, but rather reflect the subjects' personal past experiences, interpretative frame of reference for the video material, and of course brain activity not related at all to the video stimuli (e.g., olfactory information, bodily states, etc.). Subject embeddings can only do so much of the heavy lifting here. An additional limit might be imposed by the inherent variation in signal-to-noise ratio, which is for example notoriously worse \cite{olman_distortion_2009} for some of the worse predicted regions like those from the Limbic network. Further analysis will have to show how much gain there is to be made in this regard.

In conclusion, we presented a state‑of‑the‑art architecture capable of modeling fMRI responses to audio‑visual video stimuli \emph{and} of generalizing to out‑of‑distribution films. Although the model was trained only on color, English‑language episodes of \textit{Friends}, it predicts brain activity for animated shorts, stick‑figure drawings, black‑and‑white footage, and French‑language clips. These results demonstrate that VIBE captures modality-specific yet content-agnostic representations that transfer across a wide range of cinematic styles and languages.  

Looking ahead, we plan to publish a more in‑depth analysis of VIBE’s internal dynamics and feature–parcel mappings, to advance our understanding of brain function and guiding future work in neuroscience.

\section{Code availability}
The code used for this model can be found at \href{https://github.com/Bicanski-NCG/VIBE}{https://github.com/Bicanski-NCG/VIBE}, including scripts to reproduce the results in this paper.

\bibliographystyle{unsrtnat}
\bibliography{references}

\begin{thebibliography}{24}
\providecommand{\natexlab}[1]{#1}
\providecommand{\url}[1]{\texttt{#1}}
\expandafter\ifx\csname urlstyle\endcsname\relax
  \providecommand{\doi}[1]{doi: #1}\else
  \providecommand{\doi}{doi: \begingroup \urlstyle{rm}\Url}\fi

\bibitem[Yamins and DiCarlo(2016)]{yamins_using_2016}
Daniel L.~K. Yamins and James~J. DiCarlo.
\newblock Using goal-driven deep learning models to understand sensory cortex.
\newblock \emph{Nature Neuroscience}, 19\penalty0 (3):\penalty0 356--365, March 2016.
\newblock ISSN 1546-1726.
\newblock \doi{10.1038/nn.4244}.
\newblock URL \url{https://www.nature.com/articles/nn.4244}.
\newblock Publisher: Nature Publishing Group.

\bibitem[Stigliani et~al.(2017)Stigliani, Jeska, and Grill-Spector]{stigliani_encoding_2017}
Anthony Stigliani, Brianna Jeska, and Kalanit Grill-Spector.
\newblock Encoding model of temporal processing in human visual cortex.
\newblock \emph{Proceedings of the National Academy of Sciences}, 114\penalty0 (51):\penalty0 E11047--E11056, December 2017.
\newblock \doi{10.1073/pnas.1704877114}.
\newblock URL \url{https://www.pnas.org/doi/10.1073/pnas.1704877114}.
\newblock Publisher: Proceedings of the National Academy of Sciences.

\bibitem[van Gerven(2017)]{van_gerven_primer_2017}
Marcel A.~J. van Gerven.
\newblock A primer on encoding models in sensory neuroscience.
\newblock \emph{Journal of Mathematical Psychology}, 76:\penalty0 172--183, February 2017.
\newblock ISSN 0022-2496.
\newblock \doi{10.1016/j.jmp.2016.06.009}.
\newblock URL \url{https://www.sciencedirect.com/science/article/pii/S0022249616300487}.

\bibitem[Gifford et~al.(2025)Gifford, Bersch, St-Laurent, Pinsard, Boyle, Bellec, Oliva, Roig, and Cichy]{gifford_algonauts_2025}
Alessandro~T. Gifford, Domenic Bersch, Marie St-Laurent, Basile Pinsard, Julie Boyle, Lune Bellec, Aude Oliva, Gemma Roig, and Radoslaw~M. Cichy.
\newblock The {Algonauts} {Project} 2025 {Challenge}: {How} the {Human} {Brain} {Makes} {Sense} of {Multimodal} {Movies}, January 2025.
\newblock URL \url{http://arxiv.org/abs/2501.00504}.
\newblock arXiv:2501.00504 [q-bio].

\bibitem[Boyle et~al.(2020)Boyle, Pinsard, and {others}]{boyle_courtois_2020}
John~A. Boyle, Boris Pinsard, and {others}.
\newblock The {Courtois} project on neuronal modelling - 2020 data release, June 2020.
\newblock Published: Poster presented at the 2020 Annual Meeting of the Organization for Human Brain Mapping (OHBM).

\bibitem[Brett et~al.(2002)Brett, Johnsrude, and Owen]{brett_problem_2002}
Matthew Brett, Ingrid~S. Johnsrude, and Adrian~M. Owen.
\newblock The problem of functional localization in the human brain.
\newblock \emph{Nature Reviews Neuroscience}, 3\penalty0 (3):\penalty0 243--249, March 2002.
\newblock ISSN 1471-0048.
\newblock \doi{10.1038/nrn756}.
\newblock URL \url{https://doi.org/10.1038/nrn756}.

\bibitem[Schaefer et~al.(2017)Schaefer, Kong, Gordon, Laumann, Zuo, Holmes, Eickhoff, and Yeo]{schaefer_local-global_2017}
Alexander Schaefer, Ru~Kong, Evan~M Gordon, Timothy~O Laumann, Xi-Nian Zuo, Avram~J Holmes, Simon~B Eickhoff, and B~T~Thomas Yeo.
\newblock Local-{Global} {Parcellation} of the {Human} {Cerebral} {Cortex} from {Intrinsic} {Functional} {Connectivity} {MRI}.
\newblock \emph{Cerebral Cortex}, 28\penalty0 (9):\penalty0 3095--3114, July 2017.
\newblock ISSN 1047-3211.
\newblock \doi{10.1093/cercor/bhx179}.
\newblock URL \url{https://doi.org/10.1093/cercor/bhx179}.

\bibitem[Qwen et~al.(2025)Qwen, Yang, Yang, Zhang, Hui, Zheng, Yu, Li, Liu, Huang, Wei, Lin, Yang, Tu, Zhang, Yang, Yang, Zhou, Lin, Dang, Lu, Bao, Yang, Yu, Li, Xue, Zhang, Zhu, Men, Lin, Li, Tang, Xia, Ren, Ren, Fan, Su, Zhang, Wan, Liu, Cui, Zhang, and Qiu]{qwen_qwen25_2025}
Qwen, An~Yang, Baosong Yang, Beichen Zhang, Binyuan Hui, Bo~Zheng, Bowen Yu, Chengyuan Li, Dayiheng Liu, Fei Huang, Haoran Wei, Huan Lin, Jian Yang, Jianhong Tu, Jianwei Zhang, Jianxin Yang, Jiaxi Yang, Jingren Zhou, Junyang Lin, Kai Dang, Keming Lu, Keqin Bao, Kexin Yang, Le~Yu, Mei Li, Mingfeng Xue, Pei Zhang, Qin Zhu, Rui Men, Runji Lin, Tianhao Li, Tianyi Tang, Tingyu Xia, Xingzhang Ren, Xuancheng Ren, Yang Fan, Yang Su, Yichang Zhang, Yu~Wan, Yuqiong Liu, Zeyu Cui, Zhenru Zhang, and Zihan Qiu.
\newblock Qwen2.5 {Technical} {Report}, January 2025.
\newblock URL \url{http://arxiv.org/abs/2412.15115}.
\newblock arXiv:2412.15115 [cs].

\bibitem[Assran et~al.(2025)Assran, Bardes, Fan, Garrido, Howes, {Mojtaba}, {Komeili}, Muckley, Rizvi, Roberts, Sinha, Zholus, Arnaud, Gejji, Martin, Hogan, Dugas, Bojanowski, Khalidov, Labatut, Massa, Szafraniec, Krishnakumar, Li, Ma, Chandar, Meier, LeCun, Rabbat, and Ballas]{assran_v-jepa_2025}
Mido Assran, Adrien Bardes, David Fan, Quentin Garrido, Russell Howes, {Mojtaba}, {Komeili}, Matthew Muckley, Ammar Rizvi, Claire Roberts, Koustuv Sinha, Artem Zholus, Sergio Arnaud, Abha Gejji, Ada Martin, Francois~Robert Hogan, Daniel Dugas, Piotr Bojanowski, Vasil Khalidov, Patrick Labatut, Francisco Massa, Marc Szafraniec, Kapil Krishnakumar, Yong Li, Xiaodong Ma, Sarath Chandar, Franziska Meier, Yann LeCun, Michael Rabbat, and Nicolas Ballas.
\newblock V-{JEPA} 2: {Self}-{Supervised} {Video} {Models} {Enable} {Understanding}, {Prediction} and {Planning}, 2025.
\newblock URL \url{https://arxiv.org/abs/2506.09985}.

\bibitem[Chen et~al.(2023)Chen, Wu, Wang, Liu, Tompkins, Chen, Che, Yu, and Wei]{chen_beats_2023}
Sanyuan Chen, Yu~Wu, Chengyi Wang, Shujie Liu, Daniel Tompkins, Zhuo Chen, Wanxiang Che, Xiangzhan Yu, and Furu Wei.
\newblock {BEATs}: {Audio} {Pre}-{Training} with {Acoustic} {Tokenizers}.
\newblock In \emph{Proceedings of the 40th {International} {Conference} on {Machine} {Learning}}, pages 5178--5193. PMLR, July 2023.
\newblock URL \url{https://proceedings.mlr.press/v202/chen23ag.html}.
\newblock ISSN: 2640-3498.

\bibitem[Radford et~al.(2022)Radford, Kim, Xu, Brockman, McLeavey, and Sutskever]{radford_robust_2022}
Alec Radford, Jong~Wook Kim, Tao Xu, Greg Brockman, Christine McLeavey, and Ilya Sutskever.
\newblock Robust {Speech} {Recognition} via {Large}-{Scale} {Weak} {Supervision}, December 2022.
\newblock URL \url{http://arxiv.org/abs/2212.04356}.
\newblock arXiv:2212.04356 [eess].

\bibitem[Goldstein et~al.(2025)Goldstein, Wang, Niekerken, Schain, Zada, Aubrey, Sheffer, Nastase, Gazula, Singh, Rao, Choe, Kim, Doyle, Friedman, Devore, Dugan, Hassidim, Brenner, Matias, Devinsky, Flinker, and Hasson]{goldstein_unified_2025}
Ariel Goldstein, Haocheng Wang, Leonard Niekerken, Mariano Schain, Zaid Zada, Bobbi Aubrey, Tom Sheffer, Samuel~A. Nastase, Harshvardhan Gazula, Aditi Singh, Aditi Rao, Gina Choe, Catherine Kim, Werner Doyle, Daniel Friedman, Sasha Devore, Patricia Dugan, Avinatan Hassidim, Michael Brenner, Yossi Matias, Orrin Devinsky, Adeen Flinker, and Uri Hasson.
\newblock A unified acoustic-to-speech-to-language embedding space captures the neural basis of natural language processing in everyday conversations.
\newblock \emph{Nature Human Behaviour}, 9\penalty0 (5):\penalty0 1041--1055, May 2025.
\newblock ISSN 2397-3374.
\newblock \doi{10.1038/s41562-025-02105-9}.
\newblock URL \url{https://www.nature.com/articles/s41562-025-02105-9}.
\newblock Publisher: Nature Publishing Group.

\bibitem[Feichtenhofer et~al.(2019)Feichtenhofer, Fan, Malik, and He]{feichtenhofer_slowfast_2019}
Christoph Feichtenhofer, Haoqi Fan, Jitendra Malik, and Kaiming He.
\newblock {SlowFast} {Networks} for {Video} {Recognition}, October 2019.
\newblock URL \url{http://arxiv.org/abs/1812.03982}.
\newblock arXiv:1812.03982 [cs].

\bibitem[Xu et~al.(2025)Xu, Guo, He, Hu, He, Bai, Chen, Wang, Fan, Dang, Zhang, Wang, Chu, and Lin]{xu_qwen25-omni_2025}
Jin Xu, Zhifang Guo, Jinzheng He, Hangrui Hu, Ting He, Shuai Bai, Keqin Chen, Jialin Wang, Yang Fan, Kai Dang, Bin Zhang, Xiong Wang, Yunfei Chu, and Junyang Lin.
\newblock Qwen2.5-{Omni} {Technical} {Report}, March 2025.
\newblock URL \url{http://arxiv.org/abs/2503.20215}.
\newblock arXiv:2503.20215 [cs].

\bibitem[Feng et~al.(2020)Feng, Yang, Cer, Arivazhagan, and Wang]{feng_language-agnostic_2020}
Fangxiaoyu Feng, Yinfei Yang, Daniel Cer, Naveen Arivazhagan, and Wei Wang.
\newblock Language-agnostic {BERT} {Sentence} {Embedding}, 2020.
\newblock URL \url{https://arxiv.org/abs/2007.01852}.
\newblock Version Number: 2.

\bibitem[Su et~al.(2023)Su, Lu, Pan, Murtadha, Wen, and Liu]{su_roformer_2023}
Jianlin Su, Yu~Lu, Shengfeng Pan, Ahmed Murtadha, Bo~Wen, and Yunfeng Liu.
\newblock {RoFormer}: {Enhanced} {Transformer} with {Rotary} {Position} {Embedding}, November 2023.
\newblock URL \url{http://arxiv.org/abs/2104.09864}.
\newblock arXiv:2104.09864 [cs].

\bibitem[Rao and Ballard(1999)]{rao_predictive_1999}
Rajesh P.~N. Rao and Dana~H. Ballard.
\newblock Predictive coding in the visual cortex: a functional interpretation of some extra-classical receptive-field effects.
\newblock \emph{Nature Neuroscience}, 2\penalty0 (1):\penalty0 79--87, January 1999.
\newblock ISSN 1546-1726.
\newblock \doi{10.1038/4580}.
\newblock URL \url{https://www.nature.com/articles/nn0199_79}.
\newblock Publisher: Nature Publishing Group.

\bibitem[Friston and Kiebel(2009)]{friston_predictive_2009}
Karl Friston and Stefan Kiebel.
\newblock Predictive coding under the free-energy principle.
\newblock \emph{Philosophical Transactions of the Royal Society B: Biological Sciences}, 364\penalty0 (1521):\penalty0 1211--1221, May 2009.
\newblock ISSN 0962-8436.
\newblock \doi{10.1098/rstb.2008.0300}.
\newblock URL \url{https://www.ncbi.nlm.nih.gov/pmc/articles/PMC2666703/}.

\bibitem[Lindquist et~al.(2009)Lindquist, Loh, Atlas, and Wager]{lindquist_modeling_2009}
Martin~A. Lindquist, Ji~Meng Loh, Lauren~Y. Atlas, and Tor~D. Wager.
\newblock Modeling the {Hemodynamic} {Response} {Function} in {fMRI}: {Efficiency}, {Bias} and {Mis}-modeling.
\newblock \emph{Neuroimage}, 45\penalty0 (1 Suppl):\penalty0 S187--S198, March 2009.
\newblock ISSN 1053-8119.
\newblock \doi{10.1016/j.neuroimage.2008.10.065}.
\newblock URL \url{https://www.ncbi.nlm.nih.gov/pmc/articles/PMC3318970/}.

\bibitem[Thomas~Yeo et~al.(2011)Thomas~Yeo, Krienen, Sepulcre, Sabuncu, Lashkari, Hollinshead, Roffman, Smoller, Zöllei, Polimeni, Fischl, Liu, and Buckner]{thomas_yeo_organization_2011}
B.~T. Thomas~Yeo, Fenna~M. Krienen, Jorge Sepulcre, Mert~R. Sabuncu, Danial Lashkari, Marisa Hollinshead, Joshua~L. Roffman, Jordan~W. Smoller, Lilla Zöllei, Jonathan~R. Polimeni, Bruce Fischl, Hesheng Liu, and Randy~L. Buckner.
\newblock The organization of the human cerebral cortex estimated by intrinsic functional connectivity.
\newblock \emph{Journal of Neurophysiology}, 106\penalty0 (3):\penalty0 1125--1165, September 2011.
\newblock ISSN 0022-3077, 1522-1598.
\newblock \doi{10.1152/jn.00338.2011}.
\newblock URL \url{https://www.physiology.org/doi/10.1152/jn.00338.2011}.
\newblock Publisher: American Physiological Society.

\bibitem[Dixit et~al.(2025)Dixit, Fakhar, Hadaeghi, Mineault, Kording, and Hilgetag]{dixit_who_2025}
Shrey Dixit, Kayson Fakhar, Fatemeh Hadaeghi, Patrick Mineault, Konrad~P. Kording, and Claus~C. Hilgetag.
\newblock Who {Does} {What} in {Deep} {Learning}? {Multidimensional} {Game}-{Theoretic} {Attribution} of {Function} of {Neural} {Units}, June 2025.
\newblock URL \url{http://arxiv.org/abs/2506.19732}.
\newblock arXiv:2506.19732 [cs].

\bibitem[Fakhar and Dixit(2021)]{fakhar_msa_2021}
Kayson Fakhar and Shrey Dixit.
\newblock {MSA}: {A} compact {Python} package for {Multiperturbation} {Shapley} value {Analysis}., 2021.
\newblock URL \url{https://github.com/kuffmode/msa}.
\newblock Publication Title: GitHub repository.

\bibitem[Shapley and {others}(1953)]{shapley_value_1953}
Lloyd~S Shapley and {others}.
\newblock A value for n-person games.
\newblock 1953.
\newblock Publisher: Princeton University Press Princeton.

\bibitem[Olman et~al.(2009)Olman, Davachi, and Inati]{olman_distortion_2009}
Cheryl~A. Olman, Lila Davachi, and Souheil Inati.
\newblock Distortion and {Signal} {Loss} in {Medial} {Temporal} {Lobe}.
\newblock \emph{PLoS ONE}, 4\penalty0 (12):\penalty0 e8160, December 2009.
\newblock ISSN 1932-6203.
\newblock \doi{10.1371/journal.pone.0008160}.
\newblock URL \url{https://dx.plos.org/10.1371/journal.pone.0008160}.
\newblock Publisher: Public Library of Science (PLoS).

\end{thebibliography}

\end{document}